\documentclass[a4paper,twoside]{article}

\usepackage{epsfig}
\usepackage{subcaption}
\usepackage{calc}
\usepackage{amssymb}
\usepackage{amstext}
\usepackage{amsmath}
\usepackage{amsthm}
\usepackage{multicol}
\usepackage{pslatex}
\usepackage{natbib}
\usepackage{algorithm2e}
\usepackage[bottom]{footmisc}
 \usepackage{tabularray}
\usepackage{SCITEPRESS}     

\begin{document}

\title{A Review on Scientific Knowledge Extraction using Large Language Models in Biomedical Sciences}


\author{\authorname{Gabriel Lino Garcia\sup{1}\orcidAuthor{0000-0003-1236-7929}, João Renato Ribeiro Manesco\sup{1}\orcidAuthor{0000-0002-1617-5142} Pedro Henrique Paiola\sup{1}\orcidAuthor{0000-0001-9093-535X}, Lucas Miranda\sup{2}, Maria Paola de Salvo\sup{2} and João Paulo Papa\sup{1}\orcidAuthor{0000-0002-6494-7514}}
\affiliation{
\sup{1}School of Sciences, São Paulo State University (UNESP), Bauru - SP, Brazil\\
\sup{2}EasyTelling, São Paulo - SP, Brazil}
\email{\{gabriel.lino, joao.r.manesco, pedro.paiola, joao.papa\}@unesp.br\\
\{lucas, paola\}@easytelling.com}
}

\keywords{evidence synthesis, large language models, LLMs, biomedical, healthcare, literature-based discovery, knowledge extraction}

\abstract{The rapid advancement of large language models (LLMs) has opened new boundaries in the extraction and synthesis of medical knowledge, particularly within evidence synthesis. This paper reviews the state-of-the-art applications of LLMs in the biomedical domain, exploring their effectiveness in automating complex tasks such as evidence synthesis and data extraction from a biomedical corpus of documents. While LLMs demonstrate remarkable potential, significant challenges remain, including issues related to hallucinations, contextual understanding, and the ability to generalize across diverse medical tasks. We highlight critical gaps in the current research literature, particularly the need for unified benchmarks to standardize evaluations and ensure reliability in real-world applications. In addition, we propose directions for future research, emphasizing the integration of state-of-the-art techniques such as retrieval-augmented generation (RAG) to enhance LLM performance in evidence synthesis. By addressing these challenges and utilizing the strengths of LLMs, we aim to improve access to medical literature and facilitate meaningful discoveries in healthcare.}

\onecolumn \maketitle \normalsize \setcounter{footnote}{0} \vfill

\section{\uppercase{Introduction}}
\label{sec:introduction}

In recent years, the number of studies published in various fields has grown exponentially, with a wide range of applications spanning multiple disciplines~\citep{bornmann2024accelerating}. While this surge in scientific literature contributes to knowledge expansion, it poses a significant challenge for researchers and professionals attempting to stay informed of critical developments~\citep{beller2013systematic}. This trend is particularly important in fields such as medicine, where the implications of new findings can have direct and life-altering impacts on patients' lives.

However, despite the abundance of information, the practical use of many of these studies remains limited~\citep{meho2007rise}. This is often due to barriers to accessing relevant and impactful research, particularly when important findings are hidden within the vastness of the literature. The medical field exemplifies this challenge, as impactful discoveries that could improve treatment outcomes or advance medical science may go unnoticed simply because they are buried beneath a deluge of publications or presented in a highly technical manner.

Systematic reviews have long been regarded as a solution, providing comprehensive and structured analyses of the available literature. While effective, these reviews are exhaustive and frequently very technical, making them time-consuming to produce and difficult for non-experts to interpret. Consequently, they do not fully solve the issue of accessibility, especially for those who lack the specialized knowledge to extract meaningful insights from the data.

The advent of large language models (LLMs) offers a promising avenue for addressing this challenge. LLMs, powered by advancements in artificial intelligence, have the potential to assist in the navigation and synthesis of vast amounts of scientific literature. Indeed, some research has already begun to explore the potential of these models to automate aspects of the systematic review process, highlighting the possibility of their application in evidence synthesis and knowledge extraction~\citep{alshami2023harnessing}.

LLMs, such as GPT, are particularly well-suited to tasks involving knowledge extraction and summarization. They can process and synthesize vast amounts of text, distilling complex ideas into more accessible formats~\citep{ignat2023has}. This capability makes them powerful tools for navigating large datasets and filtering out the most relevant information. Moreover, the ability of LLMs to generate concise and coherent summaries from complex data sets is already being harnessed in some preliminary applications within systematic review processes.

While early applications of LLMs in systematic reviews are promising, synthesizing scientific evidence for decision-making requires more than mere automation of reviews. Managers, healthcare professionals, and other non-technical stakeholders require tools to explore the literature and synthesize evidence meaningfully and effectively. This necessitates further investigation into how LLMs can be adapted as a tool for evidence synthesis for researchers and those involved in policy-making and management in fields like healthcare.

Given the potential of LLMs to address these challenges, this paper aims to provide a comprehensive review of current research exploring the use of LLMs in evidence synthesis, with a particular focus on their application in the medical field. We will examine existing methods, tools, and frameworks that leverage LLMs for scientific knowledge extraction and discuss how these approaches can help bridge the gap between vast amounts of data and practical, actionable insights.

Thus, we have proposed a review of scientific evidence synthesis using LLMs, particularly for the healthcare case. Among the contributions of our work are:

\begin{itemize}
    \item We provide a comprehensive evaluation of the state-of-the-art LLMs used for medical evidence synthesis, focusing on their performance and challenges in tasks like clinical decision support and knowledge extraction.
    \item We identify key limitations, including issues with hallucinations, contextual understanding, and scalability, offering insights into the research gaps that need to be addressed.
    \item We propose the need for unified benchmarks and highlight future directions to enhance the effectiveness of LLMs in biomedical applications.
\end{itemize}

The remainder of this paper is structured as follows. In Section~\ref{sec.protocol}, we present the protocol used to conduct our review. Section~\ref{sec.background} gives a background on evidence synthesis based on scientific literature. Section~\ref{sec.results} discusses the main works found in scientific evidence synthesis. Section~\ref{sec.future} discusses the state of the art and current limitations in the literature, and Section~\ref{sec.conclusion} concludes the paper.

\section{Review Methodology}
\label{sec.protocol}


This systematic review followed the Preferred Reporting Items for Systematic Reviews and Meta-Analyses (PRISMA) guidelines to ensure a transparent and rigorous approach to evaluating the role of LLMs in the biomedical field. The main objective of this review is to provide a comprehensive assessment of the effectiveness, challenges, and methodologies of LLMs in the extraction and synthesis of medical knowledge. While knowledge extraction is a broader area of interest, this review specifically focuses on evidence synthesis, which aligns more closely with the complex task of integrating diverse medical information for clinical decision-making and research purposes.

The guiding research question of this review is: ``How effective are LLMs in extracting and synthesizing structured and relevant knowledge from medical texts, and what are the key challenges and outcomes associated with their application in the medical field?" This question encompasses the exploration of the models most frequently used, how they are fine-tuned for biomedical applications, their performance compared to other knowledge extraction methods, and the specific biomedical domains that benefit most from LLM-based approaches.



The review adopted a  search strategy across several academic and preprint databases, including ACM Digital Library, IEEE Xplore, Scopus, SOLO (including arXiv), and ScienceDirect. The search string was designed to cover the breadth of LLM applications in medical knowledge extraction, focusing specifically on evidence synthesis tasks. Studies were included if they focused on applying LLMs in medical settings, provided quantitative evaluations, and were published in English over the past five years.


\subsection{Screening of Relevant Works}

\begin{figure}[ht!]
    \centering
    \includegraphics[width=\linewidth]{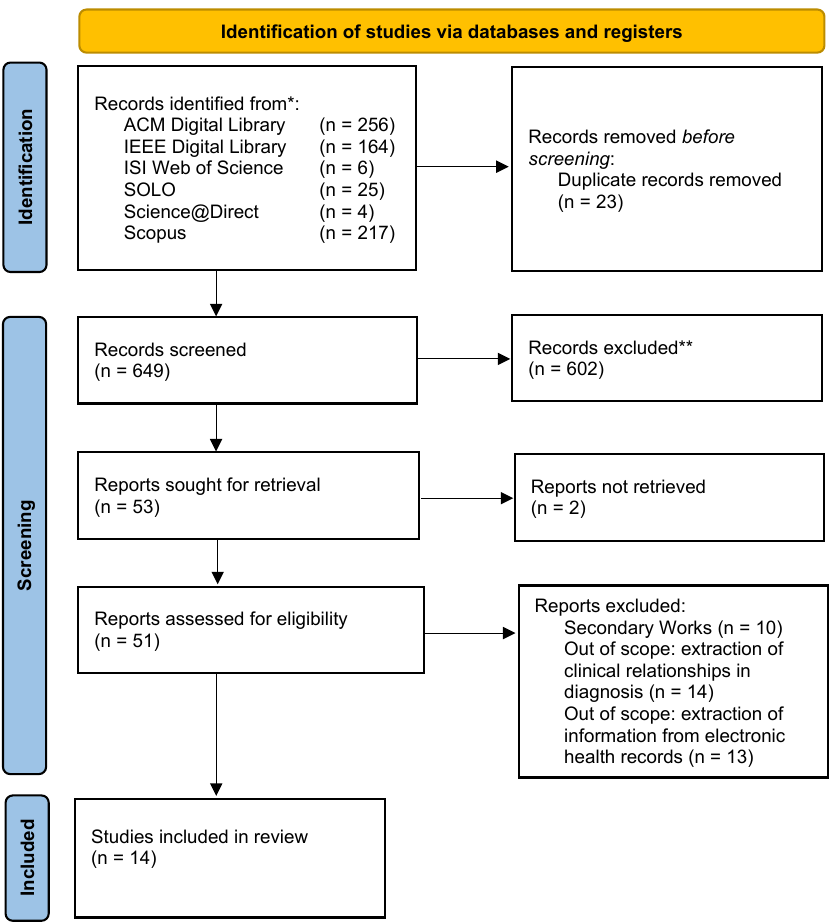}
    \caption{Illustration of the screening process, conducted in accordance with the PRISMA protocol, used to identify and select relevant studies for this review.}
    \label{fig.prisma}
\end{figure}

To enhance the screening process's efficiency, we used a ChatGPT-4 LLM to generate summaries and provide context for each selected study. This approach allowed us to hasten the review of many articles while maintaining a high level of consistency in the analysis. The initial screening was guided by predefined inclusion and exclusion criteria, with priority given to studies that addressed LLM applications in evidence synthesis. A diagram illustrating the screening and selection process, following PRISMA guidelines, is provided in Figure~\ref{fig.prisma}.

After the automated summarization phase, we manually validated the results to ensure that key insights, challenges, and contributions were accurately captured. This dual approach—combining automation with manual oversight—provided a robust method for managing the complexity of the literature while preserving the depth and rigor expected in systematic reviews.

For each included study, we extracted data on the type of LLM used, the medical task it was applied to, performance outcomes, and the challenges encountered. Particular attention was paid to how LLMs were adapted for biomedical knowledge extraction, the domains where they performed most effectively, and how they compared to other knowledge extraction techniques. 

\section{Background}
\label{sec.background}

Evidence synthesis refers to collecting, analyzing, and integrating information from the literature in a well-documented way to answer a pre-defined research question. One such form of evidence synthesis is through systematic reviews, which follow a rigorous search protocol to obtain all possible empirical evidence regarding a certain area of knowledge in a way that reduces bias and allows for replicability~\citep{lasserson2019starting}.

The objective of such systems is to support decision-makers by providing a reliable summary of research findings in the scope of the literature. However, constantly performing manual evidence synthesis whenever you need to make a decision can cause many constraints, especially since systematic reviews are highly complex and time-consuming, requiring strict adherence to rigorous protocols. This process often limits the decision maker's ability to respond quickly to novel evidence~\citep{singh2017conduct}.

Many approaches have been proposed to accelerate such processes using novel Natural Language Processing (NLP) techniques such as LLMs, mostly focusing on different aspects of the systematic review protocol, which can significantly reduce the manual burden and time required to complete a review~\citep{bolanos2024artificial}. These tools, however, are still focused on specific parts of the review process, such as screening or data extraction. While they can still be relevant, preparing the complete review can still take time and hinder decision-makers productivity.

The use of LLMs to accelerate knowledge synthesis has been explored in other reviews \citep{bolanos2024artificial,burgard2023reducing,de2023artificial}. However, these reviews primarily concentrate on streamlining the systematic review process. In contrast, our work focuses on techniques that use LLMs to perform evidence synthesis directly. In other words, those methods focus on compiling relevant literature in a more flexible, less constrained manner to provide quick, up-to-date knowledge, particularly within the healthcare domain, where timely access to synthesized evidence is crucial.

\section{Results}
\label{sec.results}
This section synthesizes the selected studies that have applied various LLMs to medical evidence synthesis. The discussion is divided into two parts: first, we examine the use of LLMs in evidence synthesis as a broader concept, and second, we focus specifically on their applications within the medical field as reported in the literature. %

\subsection{Evidence Synthesis}
Regarding evidence Synthesis using LLMs, a few works have been proposed more broadly to solve specific tasks in recent literature. One work~\citep{khraisha2024can} evaluates the efficacy of GPT-4 in performing systematic review tasks such as title/abstract screening, full-text review, and data extraction from various literature types and languages without human intervention. The study employed a "human-out-of-the-loop" approach, meaning that GPT-4 was tested independently without human intervention during extraction, focusing on the literature regarding parenting in protracted refugee situations. This work describes several challenges of current GPT models, indicating that the model's performance is heavily biased by the distribution of relevant and irrelevant studies, displaying a tendency of over-excluding studies.

In other words, this concept is expanded in less generic fields, such as government report generation~\citep{gupta2024automating}, in which Google's Gemini Pro and OpenAI's GPT-3.5 Turbo are used to improve data extraction, analysis, and visualization processes by reading all existing reports to identify graphs and charts needing updates, extracting data points from these graphs and synthesizing information in their knowledge repositories to update their report graphs. Although the method does not exactly synthesize information in textual form, it performs a Retrieval-Augmented Generation (RAG) method to synthesize information in graphic form.

The work of~\cite{yu2023bag} focuses on enhancing the extraction of training data from language models, specifically using the GPT-Neo 1.3B model in a two-stage pipeline: suffix generation, where various sampling strategies are employed, including top-k sampling, nucleus sampling, and typical sampling, to produce candidate suffixes based on a given prefix; and suffix ranking, in which perplexity and additional metrics like Zlib entropy are used to evaluate and select the most likely suffixes from the generated candidates. Although interesting for several aspects of knowledge extraction, this technique is still dependent on a large number of candidates that are ranked to identify a single successful instance, only working in limited scenarios.

In order to summarize the knowledge base, Discrete Text Summarization (DeTS) was proposed as a novel unsupervised method for discrete text summarization, in which grammatical sequence scoring and independent key points from large textual datasets are used for natural language inference. The technique consists of two main components: candidate extraction and candidate matching. In the candidate extraction phase, the algorithm utilizes Semantic Role Labeling (SRL) to segment comments into smaller pieces. The candidate matching phase aligns these key points with sentences from the entire corpus using NLI algorithms, determining whether the sentences entail the key points based on their semantic similarity. Although the method was successful in a restricted scenario of anonymous comments, there is still a concern regarding bias in the employed dataset.

Focusing more on the extraction of information in scientific texts,~\cite{dagdelen2024structured} focuses on joint-named entity recognition and relation extraction, allowing the models to identify entities and their relationships within materials science literature. The model architecture is based on a sequence-to-sequence framework, where GPT-3 and LLaMa-2 models are fine-tuned on a relatively small dataset of 400-650 annotated text-extraction pairs. Given the dataset's limitations and the relationships' complexity, the model still struggles to perform complete evidence synthesis.


\subsection{Synthesis in Biomedical Sciences}

In the healthcare domain, which is the primary focus of our analysis, one study explores the application of LLMs, specifically GPT-3.5 and GPT-4, for Literature-Based Discovery to generate research hypotheses~\cite{nedbaylo2024implementing}. The authors employ a bifurcated prompt engineering technique, where the original prompt is split into two parts and executed in separate chat windows. In the first segment, the model generates a list of disease characteristics without disclosing the disease name. In contrast, the second suggests potential interventions based on the factors identified in the initial segment. The study conducts a qualitative evaluation and concludes that GPT-based methods heavily depend on preexisting bibliographic databases, such as Medline, which restricts their applicability to fields with well-established datasets. Additionally, the study finds that outputs from GPT-4 often lack technical depth and novelty, and they frequently fail to provide adequate references for the generated information.

\cite{tao2024gpt} employs the OpenAI GPT-4 model to evaluate its performance in answering specific questions related to HIV drug resistance from published scientific papers. The researchers designed an automated pipeline that transformed the text of 60 selected HIV drug resistance papers into markdown format, excluding sections like the introduction and discussion to focus on the methods and results. By posing 60 questions in two modes, multiple-question mode (all questions presented simultaneously) and single-question mode (one question at a time), with and without an instruction sheet containing specialized knowledge, the authors were able to perform a quantitative evaluation of the task. Although the method obtained a fairly good mean accuracy of $86.9\%$, some aspects of the technique could be improved, such as the fact that the instruction sheet was not effectively utilized by the model, by not improving the accuracy of the responses, in addition, the method struggled with specific queries that required making inferences, especially when dealing with implicit information in the texts. Worst of all, the GPT-4 model was more likely to provide false positive answers when questions were submitted individually than when they were submitted together, indicating the presence of bias when dealing with certain aspects of the literature.

Using the Claude 2 LLM, one study performs a proof-of-concept design to evaluate the performance of LLMs in interpreting randomized controlled trials (RCTs)~\cite{gartlehner2024data}. To do that, the researchers selected a convenience sample of 10 open-access RCTs and focused on extracting 16 distinct data elements, which included study identifiers, participant characteristics, and outcome data. The data extraction pipeline involves mostly prompt engineering, where the crafted prompts were iteratively tested and refined to optimize the model's output. Although the method obtains interesting results, several concerns should be considered, such as the fact that open RCTs were used, meaning they could be in the training set of Claude 2. The authors also reported two cases of hallucinations from answers that the model did not know how to reply to.

Focusing on information retrieval to support knowledge synthesis in biomedical documents, one work uses the RoBERTa language model to work with the COVID-19 Open Research Dataset~\cite{saxena2022large}. To do that, an architecture was developed consisting of three main components that extract relevant information:  Paragraph Retrieval, Triplet Retrieval from Knowledge Graphs, and Complex Question Answering. The Paragraph Retrieval employs a hybrid approach combining lexical methods with semantic search for indexing and re-ranking results based on relevance. The Triplet Retrieval system extracts subject-relation-object triplets from the knowledge graph, allowing for faceted refinement of search results. Finally, the complex QA system utilizes a Multi-hop Dense Retriever to handle multi-hop questions, iteratively retrieving relevant passages and employing both extractive (RoBERTa) and generative (Fusion-in-Decoder) readers to generate answers. The model relies on an older approach for language models, which, although smaller, has more difficulty when dealing with generative language answers; the authors also report difficulty in dealing with the amount of unstructured textual data reported in the medical literature.

This trend of using encoder-based smaller language models was also observed in two other works in the literature. One of them works on a dataset of biomedical articles collected from PubMed, aiming to summarize biomedical literature by using an encoder-decoder model based on BioBERT~\cite{alambo2022entity} To achieve its purpose, the researchers created a framework based on two main components: an entity-driven knowledge retriever that extracts facts based on named entities identified in the source documents, and a knowledge-guided abstractive summarizer that generates summaries by attending to both the source article and the retrieved facts.

The other work focuses on semi-automate data extraction for systematic literature using a BERT model pre-trained on a corpus of 100,000 open-access clinical publications, the main objective being extracting clinical relationships from the corpus in order to perform literature filtering~\cite{panayi2023evaluation}. Both works, although displaying interesting results, suffer from the difficulties of dealing with complex unstructured data; the BioBERT-based models, although displaying better results, still report hallucination, as there is an attempt to generate structured text.

Other works are more focused on extracting specific information from the biomedical literature. One such work, named ChIP-GPT, fine-tuned a LLaMa architecture on Sequence Read Archive information obtained from biomedical database records by extracting relevant sentences and summarizing biomedical records, intending to identify chromatin immunoprecipitation (ChIP) targets and cell lines in these records. A summarization step was implemented to manage the input length limitations of the model for longer records. This involved selecting informative sentences while discarding irrelevant details, ensuring the model could generate concise and accurate answers. The final results display good accuracy in the task even when compared with human experts. However, it was observed that this summarization step still had a problem discarding relevant information for the model, which impacted the final results. 

Following the same concept, another work focuses on extracting relevant data from scientific literature for chemical risk assessment, focusing on Bisphenol A~\cite{sonnenburg2024artificial}. To achieve that, the authors created a dataset containing 78 selected publications, primarily extracting results sections to form prompt-completion pairs so that they could perform fine-tuning on an LLM from the GPT-3 family named Curie model. The Curie model reported remarkable results, especially compared to other ready-to-use models that were not fine-tuned, even if the authors still have some concerns regarding treating studies with more complex experimental designs.

Finally, a study that focuses on using a Retriever-Augmented Generation (RAG) model to extract and deliver accurate medical knowledge about diabetes and diabetic foot care, specifically tailored for laypersons with an eighth-grade literacy level, was proposed~\cite{mashatian2024building}. The authors use the GPT-4 model to formulate user-friendly responses based on the retrieved context, offering a zero-shot solution to the problem. A set of 295 articles was used to provide context and be evaluated across 175 questions on various diabetes-related topics. This work offers an interesting solution that does not require training the LLM and displays the results in natural language, even for people with lower literacy levels. Despite its advantages, the model is still evaluated on a very limited set of data, so further experiments are required to see how it performs with new works being published in the literature.

\section{Comparison of Current Methods and Future Directions}
\label{sec.future}

In this section, we perform an aggregated analysis that compares datasets, performance metrics, identified limitations, and future trends suggested by each method to understand the limitations of current techniques and the future trends in the literature. Although the area lacks a complete benchmark for evaluating evidence synthesis, and each method does a different analysis, we still compare each technique by their main points and context in Table~\ref{tab.comparison}.


\begin{table*}[h!]
\centering
\caption{Comparison of various methods used in biomedical literature for knowledge extraction and evidence synthesis.}
\label{tab.comparison}
\resizebox{\textwidth}{!}{%
\begin{tblr}{
  width = 1.7\textwidth,
  colspec = {Q[200]Q[100]Q[150]Q[290]Q[150]Q[370]},
  row{1} = {c},
  cell{2}{2} = {c},
  cell{2}{3} = {c},
  cell{3}{2} = {c},
  cell{3}{3} = {c},
  cell{4}{2} = {c},
  cell{4}{3} = {c},
  cell{5}{2} = {c},
  cell{5}{3} = {c},
  cell{6}{2} = {c},
  cell{6}{3} = {c},
  cell{7}{2} = {c},
  cell{7}{3} = {c},
  cell{8}{2} = {c},
  cell{8}{3} = {c},
  cell{9}{2} = {c},
  cell{9}{3} = {c},
  cell{10}{2} = {c},
  cell{10}{3} = {c},
  hlines,
  vline{2-6} = {-}{},
}
\textbf{Paper}                                   & \textbf{Year} & \textbf{Model}    & \textbf{Biomedical Context}                                                                                                                                                                 & \textbf{Reported Results}                                          & \textbf{Brief Summary}                                                                                                                                                                                                                                                   \\
\cite{alambo2022entity}         & 2022          & BioBERT           & Biomedical articles and facts from biomedical knowledge bases.                                                                                                                                   & Precision: 55.07\%, Recall 7.97\%                         & The method uses integrates named entities and facts from biomedical knowledge bases into transformer-based models to enhance the factual consistency of generated summaries in biomedical literature.                                                                    \\
\cite{saxena2022large}          & 2022          & RoBERTa + Fusion-in-Decoder           & Unstructured text data from documents, articles, biomedical journals, and structured data like tables and electronic health records.                                                             & Precision: 81\%, NDCG: 72\%                               & A framework for complex information retrieval from biomedical documents is developed, utilizing paragraph retrieval, triplet retrieval from knowledge graphs, and complex question answering.                    \\
\cite{panayi2023evaluation}     & 2023          & BERT              & Metrics for progression-free survival, patient age, treatment arm dosage, and eGFR measurements extracted from~systematic literature reviews in oncology and Fabry disease.                      & F1-score:~73\%                                            & A machine learning approach using pre-trained BERT and CRF models was employed to automate data extraction from systematic literature reviews by identifying biomedical entities and their relations.                                                                    \\
\cite{sonnenburg2024artificial} & 2024          & GPT-3             & NAM-based toxicity data on Bisphenol A (BPA) extracted from scientific publications                                                                                                              & Precision: 25\%, Recall: 23\%, F1-score: 50\%.            & A fine-tuned version of GPT-3 was employed to extract and structure data from scientific publications for risk assessment, specifically targeting toxicological properties of bisphenol A (BPA).                                                                         \\
\cite{nedbaylo2024implementing} & 2024          & GPT-3.5 and GPT-4 & Scientific papers and literature related to biomedical concepts                                                                                                                                  & Only a qualitative evaluation was performed in this paper & A Literature-based discovery technique based on the generation of research hypotheses via bifurcated prompt engineering.                                                                                                                                                 \\
\cite{cinquin2024chip}          & 2024          & ChIP-GPT (LLaMa)  & Metadata from Chromatin Immunoprecipitation Sequencing (ChIP-Seq) experiments.                                                                                                                   & Accuracy: 90\%                                            & A LLaMA-based model is fine-tuned to extract metadata from biomedical database records, targeting chromatin immunoprecipitation (ChIP) data.                                                                                   \\
\cite{gartlehner2024data}       & 2024          & Claude 2          & The research paper focuses on extracting several types of biomedical data elements from published studies, including study identifiers, characteristics of study participants, and outcome data. & Accuracy: 96.3\%, F1-score: 98\%                          & The study uses the Claude 2 LLM to extract several types of biomedical elements from published studies in medical literature.                                                                                                                                            \\
\cite{mashatian2024building}    & 2024          & GPT-4             & Scientific papers related to diabetes and diabetic foot care.                                                                                                                                    & Accuracy: 98\%                                            & A RAG model was developed to extract accurate medical knowledge about diabetes and diabetic foot care for laypersons with an eighth-grade literacy level, using prompt engineering. \\
\cite{tao2024gpt}               & 2024          & GPT-4             & HIV genetic sequence data, antiviral treatment histories, and the effects of mutations on susceptibility to antiviral drugs from papers on HIV drug resistance (HIVDR).                          & Accuracy: 86.9\%, Recall 72.5\%, Precision 87.4\%         & An automated pipeline using GPT-4 was created to evaluate the accuracy of responses to queries about published papers on HIV drug resistance.                                                      
\end{tblr}}
\end{table*}

Existing methods that use language models for knowledge synthesis in biomedical tasks employ various strategies and applications, addressing several aspects of knowledge extraction, including literature-based discovery, data extraction, and question-answering. One of the key challenges in this area is to perform a comprehensive evaluation capable of comparing the effectiveness of each method and LLM by themselves. Currently, each method follows its own protocol, with distinct datasets, evaluation criteria, and tasks, making it difficult to draw direct comparisons between models or assess their generalizability to new problems.

In the reviewed studies, several models have been used for evidence syntheses, such as GPT-3.5, GPT-4, Claude 2, and even encoder-based methods, such as RoBERTa and BioBERT, on a wide range of tasks ranging from generating research hypotheses to answering specific scientific queries and extracting clinical trial data. For example, \cite{nedbaylo2024implementing} focuses on literature-based discovery through a GPT-4 LLM, while \cite{tao2024gpt} employs the same LLM in a very distinct task and context: processing structured biomedical data for HIV drug resistance analysis. Although both methods use the same model and follow similar strategies, it is difficult to conclude if the observed limitations come from the technique or the contextual data, as there is no shared information among them. Both studies also highlight considerable limitations of current LLMs in dealing with insights from the data, especially when the model needs to handle implicit information to generate connections and process novel information.

Another current problem in the literature is that most methods focus on using pre-established datasets, such as Medline in the study by \cite{nedbaylo2024implementing} or the curated HIV drug resistance literature in \cite{tao2024gpt}, to create a knowledge base for the LLM and fine-tune the model. In this case, it is difficult to establish the influence of each piece of data in the response, and it also makes dealing with the novel and growing literature even more difficult, as there is a need to re-train the model. As such, strategies based on RAG, such as the one proposed by \cite{mashatian2024building}, appear as a more reliable solution to deal with new pieces of data and may pose the path forward for new techniques that may incorporate modern RAG approaches at their core.

In the study of  \cite{gartlehner2024data} that uses a Claude 2 model for the extraction of data from RCTs, the authors raise two main concerns regarding this area: the growing risks of hallucination caused by difficulties in dealing with unstructured data in biomedical research, and the potential bias coming from training on open-access data, which the model may already have seen during training. In addition, without a common benchmark, it is difficult to determine how these models would fare on more diverse datasets or in more complex, less predictable environments.

Studies using smaller, encoder-based models like RoBERTa, BioBERT, and BERT also exhibit this challenge. While they are effective for specific tasks like entity extraction and information retrieval, their scalability and ability to generate new insights are limited. This is particularly evident in models designed for more specialized tasks, such as ChIP-GPT for chromatin immunoprecipitation data extraction and the fine-tuned Curie model for chemical risk assessment. These models achieve impressive results within their respective domains but lack the flexibility to be easily compared or adapted across different biomedical tasks without a unified evaluation standard.

The absence of a unified benchmark also extends to the issue of hallucinations and false positives, which have been identified across multiple studies. For example, the GPT-4 model in \cite{tao2024gpt}'s study showed a higher tendency for false positives in single-question mode, raising concerns about the consistency and reliability of LLMs when applied to individual biomedical queries. Likewise, the hallucinations observed in Claude 2's responses in \cite{gartlehner2024data}'s work underlines the need for more robust evaluation methods to assess generated content's accuracy and factual correctness.

As we can observe from the current state of the field, developing a unified benchmark to standardize the evaluation of LLMs across evidence synthesis in diverse biomedical tasks is crucial to advance the field. This benchmark must evaluate the model's accuracy, contextual comprehension, insights capability, generalizability, and the capacity to process unstructured data commonly found in this area. Furthermore, recent advancements in LLM retrieval techniques, such as RAG techniques, could significantly enhance evidence synthesis, as these techniques not only improve the quality of generated responses but also allow for the models to deal with the novel incoming literature, on top of enabling the retrieval of the relevant sources, ensuring that LLMs contribute to evidence synthesis in a transparent and verifiable manner.

\section{Conclusion}
\label{sec.conclusion}

This review aims to highlight the increasing role of natural language processing  and LLMs in biomedical research, particularly for automating tasks like literature-based discovery, clinical trial data extraction, and answering complex scientific questions. Although these models show great promise, they also reveal significant shortcomings. Many rely on preexisting datasets and struggle with more nuanced tasks such as making inferences, understanding context, and dealing with implicit information. Issues like hallucinations and false positives in several studies further underscore the need for stronger fact-checking and better handling of contextual information.

A major challenge is the lack of a unified benchmark across various tasks. The current variety of datasets and evaluation metrics makes it difficult to compare models directly or evaluate how well they generalize to new problems. Fine-tuned models like ChIP-GPT and Curie perform well in narrow areas but are less proven in broader contexts. Similarly, older encoder-based models such as RoBERTa and BioBERT have scalability limitations and are difficult to manage the complexity of unstructured medical literature.

Looking ahead, efforts should focus on improving models' ability to grasp context, addressing the problem of hallucinations, and creating unified benchmarks to standardize evaluation across tasks. Integrating techniques like knowledge graphs, multi-hop reasoning, and retrieval-augmented generation (RAG) could help close the current performance gaps, especially for complex evidence synthesis. While LLMs hold great potential in healthcare, they still require improvements in architecture, evaluation methods, and unstructured data handling to fully realize their transformative potential in biomedical research and automated evidence synthesis.

\section*{\uppercase{Acknowledgements}}
This study was funded by the S\~ao Paulo Research Foundation (FAPESP) grants $2013/07375-0$, $2019/07665-4$, $2023/14427-8$, $2024/00789-8$, and $2024/01336-7$ as well as the National Council for Scientific and Technological Development 
 grants $308529/2021-9$ and $400756/2024-2$.

\bibliographystyle{apalike}
{\small
\bibliography{example}}

\begin{thebibliography}{}

\bibitem[Alambo et~al., 2022]{alambo2022entity}
Alambo, A., Banerjee, T., Thirunarayan, K., and Raymer, M. (2022).
\newblock Entity-driven fact-aware abstractive summarization of biomedical
  literature.
\newblock In {\em 2022 26th International Conference on Pattern Recognition
  (ICPR)}, pages 613--620. IEEE.

\bibitem[Alshami et~al., 2023]{alshami2023harnessing}
Alshami, A., Elsayed, M., Ali, E., Eltoukhy, A.~E., and Zayed, T. (2023).
\newblock Harnessing the power of chatgpt for automating systematic review
  process: Methodology, case study, limitations, and future directions.
\newblock {\em Systems}, 11(7):351.

\bibitem[Beller et~al., 2013]{beller2013systematic}
Beller, E.~M., Chen, J. K.-H., Wang, U. L.-H., and Glasziou, P.~P. (2013).
\newblock Are systematic reviews up-to-date at the time of publication?
\newblock {\em Systematic reviews}, 2:1--6.

\bibitem[Bolanos et~al., 2024]{bolanos2024artificial}
Bolanos, F., Salatino, A., Osborne, F., and Motta, E. (2024).
\newblock Artificial intelligence for literature reviews: Opportunities and
  challenges.
\newblock {\em arXiv preprint arXiv:2402.08565}.

\bibitem[Bornmann et~al., 2024]{bornmann2024accelerating}
Bornmann, L., Haunschild, R., and Mutz, R. (2024).
\newblock Accelerating scientific progress with preprints.
\newblock {\em Nature Computational Science}, 4(5):311--311.

\bibitem[Burgard and Bittermann, 2023]{burgard2023reducing}
Burgard, T. and Bittermann, A. (2023).
\newblock Reducing literature screening workload with machine learning.
\newblock {\em Zeitschrift f{\"u}r Psychologie}.

\bibitem[Cinquin, 2024]{cinquin2024chip}
Cinquin, O. (2024).
\newblock Chip-gpt: a managed large language model for robust data extraction
  from biomedical database records.
\newblock {\em Briefings in bioinformatics}, 25(2):bbad535.

\bibitem[Dagdelen et~al., 2024]{dagdelen2024structured}
Dagdelen, J., Dunn, A., Lee, S., Walker, N., Rosen, A.~S., Ceder, G., Persson,
  K.~A., and Jain, A. (2024).
\newblock Structured information extraction from scientific text with large
  language models.
\newblock {\em Nature Communications}, 15(1):1418.

\bibitem[de~la Torre-L{\'o}pez et~al., 2023]{de2023artificial}
de~la Torre-L{\'o}pez, J., Ram{\'\i}rez, A., and Romero, J.~R. (2023).
\newblock Artificial intelligence to automate the systematic review of
  scientific literature.
\newblock {\em Computing}, 105(10):2171--2194.

\bibitem[Gartlehner et~al., 2024]{gartlehner2024data}
Gartlehner, G., Kahwati, L., Hilscher, R., Thomas, I., Kugley, S., Crotty, K.,
  Viswanathan, M., Nussbaumer-Streit, B., Booth, G., Erskine, N., et~al.
  (2024).
\newblock Data extraction for evidence synthesis using a large language model:
  A proof-of-concept study.
\newblock {\em Research Synthesis Methods}.

\bibitem[Gupta et~al., 2024]{gupta2024automating}
Gupta, R., Pandey, G., and Pal, S.~K. (2024).
\newblock Automating government report generation: A generative ai approach for
  efficient data extraction, analysis, and visualization.
\newblock {\em Digital Government: Research and Practice}.

\bibitem[Ignat et~al., 2023]{ignat2023has}
Ignat, O., Jin, Z., Abzaliev, A., Biester, L., Castro, S., Deng, N., Gao, X.,
  Gunal, A., He, J., Kazemi, A., et~al. (2023).
\newblock Has it all been solved? open nlp research questions not solved by
  large language models.
\newblock {\em arXiv preprint arXiv:2305.12544}.

\bibitem[Khraisha et~al., 2024]{khraisha2024can}
Khraisha, Q., Put, S., Kappenberg, J., Warraitch, A., and Hadfield, K. (2024).
\newblock Can large language models replace humans in systematic reviews?
  evaluating gpt-4's efficacy in screening and extracting data from
  peer-reviewed and grey literature in multiple languages.
\newblock {\em Research Synthesis Methods}.

\bibitem[Lasserson et~al., 2019]{lasserson2019starting}
Lasserson, T.~J., Thomas, J., and Higgins, J.~P. (2019).
\newblock Starting a review.
\newblock {\em Cochrane handbook for systematic reviews of interventions},
  pages 1--12.

\bibitem[Mashatian et~al., 2024]{mashatian2024building}
Mashatian, S., Armstrong, D.~G., Ritter, A., Robbins, J., Aziz, S., Alenabi,
  I., Huo, M., Anand, T., and Tavakolian, K. (2024).
\newblock Building trustworthy generative artificial intelligence for diabetes
  care and limb preservation: A medical knowledge extraction case.
\newblock {\em Journal of Diabetes Science and Technology}, page
  19322968241253568.

\bibitem[Meho, 2007]{meho2007rise}
Meho, L.~I. (2007).
\newblock The rise and rise of citation analysis.
\newblock {\em Physics World}, 20(1):32.

\bibitem[Nedbaylo and Hristovski, 2024]{nedbaylo2024implementing}
Nedbaylo, A. and Hristovski, D. (2024).
\newblock Implementing literature-based discovery (lbd) with chatgpt.
\newblock In {\em 2024 47th MIPRO ICT and Electronics Convention (MIPRO)},
  pages 120--125. IEEE.

\bibitem[Panayi et~al., 2023]{panayi2023evaluation}
Panayi, A., Ward, K., Benhadji-Schaff, A., Ibanez-Lopez, A.~S., Xia, A., and
  Barzilay, R. (2023).
\newblock Evaluation of a prototype machine learning tool to semi-automate data
  extraction for systematic literature reviews.
\newblock {\em Systematic Reviews}, 12(1):187.

\bibitem[Saxena et~al., 2022]{saxena2022large}
Saxena, S., Sangani, R., Prasad, S., Kumar, S., Athale, M., Awhad, R., and
  Vaddina, V. (2022).
\newblock Large-scale knowledge synthesis and complex information retrieval
  from biomedical documents.
\newblock In {\em 2022 IEEE International Conference on Big Data (Big Data)},
  pages 2364--2369. IEEE.

\bibitem[Singh, 2017]{singh2017conduct}
Singh, S. (2017).
\newblock How to conduct and interpret systematic reviews and meta-analyses.
\newblock {\em Clinical and translational gastroenterology}, 8(5):e93.

\bibitem[Sonnenburg et~al., 2024]{sonnenburg2024artificial}
Sonnenburg, A., van~der Lugt, B., Rehn, J., Wittkowski, P., Bech, K., Padberg,
  F., Eleftheriadou, D., Dobrikov, T., Bouwmeester, H., Mereu, C., et~al.
  (2024).
\newblock Artificial intelligence-based data extraction for next generation
  risk assessment: Is fine-tuning of a large language model worth the effort?
\newblock {\em Toxicology}, 508:153933.

\bibitem[Tao et~al., 2024]{tao2024gpt}
Tao, K., Osman, Z.~A., Tzou, P.~L., Rhee, S.-Y., Ahluwalia, V., and Shafer,
  R.~W. (2024).
\newblock Gpt-4 performance on querying scientific publications:
  reproducibility, accuracy, and impact of an instruction sheet.
\newblock {\em BMC Medical Research Methodology}, 24(1):139.

\bibitem[Yu et~al., 2023]{yu2023bag}
Yu, W., Pang, T., Liu, Q., Du, C., Kang, B., Huang, Y., Lin, M., and Yan, S.
  (2023).
\newblock Bag of tricks for training data extraction from language models.
\newblock In {\em International Conference on Machine Learning}, pages
  40306--40320. PMLR.

\end{thebibliography}

\end{document}